\documentclass[twoside]{article}

\usepackage[accepted]{aistats2022}
\usepackage[round]{natbib}

\usepackage[colorlinks=true, allcolors=blue]{hyperref}
\usepackage{booktabs}
\usepackage{url}
\usepackage{amsmath}
\usepackage{amsfonts}
\usepackage{amsthm}
\usepackage{graphicx}
\usepackage{amsmath, bm}
\usepackage{amsbsy}
\usepackage{dsfont}
\usepackage{algpseudocode}
\usepackage{tabularx}
\usepackage{wrapfig}
\usepackage{caption}
\usepackage{subcaption}
\usepackage{adjustbox}

\newcommand*\mean[1]{\bar{#1}}
\def \N {\mathcal{N}}
\def \GP {\operatorname{GP}}
\def \PG {\operatorname{PG}}
\def \cosh {\operatorname{cosh}}
\def \Po {\operatorname{Po}}
\def \diag {\operatorname{diag}}
\def \co {\operatorname{Const}}
\def \ga {\operatorname{Ga}}
\def \cat {\operatorname{Cat}}
\def \dir {\operatorname{Dir}}
\newcommand{\mname}{FABLE\xspace}

\newcommand\best[1]{\textcolor{red}{#1}}
\newcommand\secbest[1]{\textcolor{blue}{#1}}

\usepackage{amsmath,amsfonts,bm}
\usepackage{xspace}

\newcommand{\etc}{\emph{etc.}\xspace} 
\newcommand{\ie}{\emph{i.e.}\xspace} 
\newcommand{\eg}{\emph{e.g.\xspace}} 

\newcommand{\nop}[1]{}

\def\eqref#1{equation~\ref{#1}}

\def\1{\bm{1}}

\DeclareMathAlphabet{\mathsfit}{\encodingdefault}{\sfdefault}{m}{sl}
\SetMathAlphabet{\mathsfit}{bold}{\encodingdefault}{\sfdefault}{bx}{n}

\newcommand{\E}{\mathbb{E}}

\DeclareMathOperator*{\argmax}{arg\,max}

\runningtitle{Leveraging Instance Features for Label Aggregation in Programmatic Weak Supervision}
\begin{document}
\twocolumn[
\aistatstitle{Leveraging Instance Features for Label Aggregation \\in Programmatic Weak Supervision}
\aistatsauthor{ Jieyu Zhang* \And Linxin Song* \And  Alexander Ratner }
\aistatsaddress{ University of Washington \And  Waseda University \And University of Washington } 
]

\begin{abstract}
Programmatic Weak Supervision (PWS) has emerged as a widespread paradigm to synthesize training labels efficiently. The core component of PWS is the \emph{label model}, which infers true labels by aggregating the outputs of multiple noisy supervision sources abstracted as \emph{labeling functions (LFs)}. Existing statistical label models typically rely only on the outputs of LF, ignoring the instance features when modeling the underlying generative process. In this paper, we attempt to incorporate the instance features into a statistical label model via the proposed \mname.
In particular, it is built on a mixture of Bayesian label models, each corresponding to a global pattern of correlation, and the coefficients of the mixture components are predicted by a Gaussian Process classifier based on instance features.
We adopt an auxiliary variable-based variational inference algorithm to tackle the non-conjugate issue between the Gaussian Process and Bayesian label models.
Extensive empirical comparison on eleven benchmark datasets sees \mname achieving the highest averaged performance across nine baselines. 
Our implementation of \mname can be found in \url{https://github.com/JieyuZ2/wrench/blob/main/wrench/labelmodel/fable.py}.
\end{abstract}

\section{INTRODUCTION}

The deployment of machine learning models typically relies on large-scale labeled data to regularly train and evaluate the models.
To collect labels, practitioners have increasingly resorted to Programmatic Weak Supervision (PWS)~\citep{ratner2016data, zhang2022survey}, a paradigm in which labels are generated cheaply and efficiently.
Specifically, in PWS, users develop weak supervision sources abstracted as simple programs called \emph{labeling functions (LFs)}, rather than make individual annotations.
These LFs could efficiently produce noisy votes on the true label or abstain from voting based on external knowledge bases, heuristic rules, \etc.
To infer the true labels, various \emph{statistical label models}~\citep{ratner2016data, ratner2019training, fu2020fast} are developed to aggregate the labels output by LFs.

One of the major technical challenges in PWS is how to  infer the true labels given the noisy and potentially conflict labels of multiple LFs. 
While being diverse in assumptions and modeling techniques, existing statistic label models typically rely solely on the LFs' labels~\citep{ratner2016data, bach2017learning, varma2017inferring, cachay2021dependency}. In this paper, we argue that incorporating instance features into a statistical label model has significant potential to improve the inferred truth.
Intuitively, statistical label models aim to recover the pattern of correlation between the LF labels and the ground truth; it is natural to assume that similar instants would share a similar pattern and therefore the instance features could be indicative of the pattern of each instant. 
When ignoring the instance features, statistical label models have to assume that the patterns or the LF correctness is instant-independent, which is unlikely to be true for real-world dataset.

To attack this problem, we propose \mname (Feature-Aware laBeL modEl), which exploits the instance features to help identify the correlation pattern of instants. We build \mname upon a recent model named EBCC~\citep{li2019exploiting}, which is a mixture model where each mixture component is a popular Bayesian extension to the DS model~\citep{dawid1979maximum} and aims to capture one sort of LF and true label correlation. To incorporate instance features, we propose to make the mixture coefficients a categorical distribution explicitly depending on instance features. In particular, a predictive Gaussian process (GP) is adopted to learn the distribution of mixture coefficients, connecting the correlation patterns with instance features.
However, the categorical distribution of mixture coefficients is non-conjugate to the Gaussian prior, hindering the usage of efficient Bayesian inference algorithm, \eg, variational inference. To overcome this, we introduce a number of auxiliary variables to augment the likelihood function to achieve the desired conjugate representation of our model. Note that there are a couple of recently proposed neural network-based models~\citep{ren2020denoising, ruhling2021end} that also leverage instance features, but via neural network. We include them as baselines for comparison and highlight that these neural network-based models typically require gold validation set for hyperparameter tuning and early stopping to be performant with comparison to a statistical model like \mname.

We conduct extensive experiments on synthetic dataset with varying size and 11 benchmark datasets. Compared with state-of-the-art baselines, \mname achieves the highest averaged performance and ranking. 
More importantly, to help understand when \mname works well and verify our arguments, we measure the correlation of instance features and the LF correctness, \ie, Corr(X, LFs). 
Then, we calculate the Pearson’s correlation coefficient between Corr(X, LFs) and the gain of \mname over EBCC on synthetic dataset, which is 0.496 with $p$-value $<0.01$, indicating that leveraging instance feature is more beneficial when the LF correctness indeed depends on the features.
\section{RELATED WORKS}
In PWS, researches have developed a bunch of statistical label model.
\citet{ratner2016data} models the joint distribution between LF and ground truth labels to describe the distribution in terms of pre-defined factor functions.
\citet{ratner2019training} models the distribution via a Markov network and recover the parameters via a matrix completion-style approach, while \citet{fu2020fast} models the distribution via a binary Ising model and recover the parameters by triplet methods.  There are other statistical models designed for extended PWS setting~\citep{shin2021universalizing} or for extended definition of LFs, \eg, partial labeling functions~\citep{yu2022learning}, indirect labeling functions~\citep{zhang2021creating}, and positive-only labeling functions~\citep{zhang2022binary}.
Besides the statistical label models, researchers have recently proposed neural network-based models to leverage instance features~\citep{ren2020denoising, ruhling2021end}, while in this work, we aim to incorporate instance features into a pure statistical model.

Prior to PWS, statistical models for label aggregation were separately developed in the field of crowdsourcing.
\citet{dawid1979maximum} used a confusion matrix parameter to generative model LF labels conditioned on the item's true annotation, for clinical diagnostics.
\citet{kim2012ibcc} formulated a Bayesian generalization with Dirichlet priors and inference by Gibbs sampling, while \citet{li2019exploiting} incorporate the subtypes as mixture correlation and decoupled the confusion matrix, which make the Bayesian generation process become a mixture model.
Their analysis of inferred worker confusion matrix clustering is a natural precursor to modelling worker correlation.

\section{PRELIMINARIES}
In this section, we first introduce the setup and notation of the programmatic weak supervision (PWS), then discuss two representative Bayesian models that can be used in PWS. We also discuss the multi-class Gaussian process classification, which is related to our proposed method.

\subsection{Notation}
Let $X=\{\vec{x}_1,...,\vec{x}_N\}$ denote a training set with $N$ featured data samples.
Assume that there are $L$ labeling functions (LFs) $\vec{y}_i=[y_{i1},...,y_{iL}]$ with $j\in[L]$, each of which classifies $N$ each sample into one of $K$ categories or abstain (outputting $-1$). 
Let $z_i$ be the latent true label of the sample $i$, $y_{ij}$ the label that LF $j$ assigns to the item $i$, $Y_i$ the set of LFs who have labelled the item $i$. 

\subsection{Bayesian Classifier Combination (BCC) Models}
\paragraph{Independent BCC.}
The iBCC~\citep{kim2012ibcc} model is a directed graphical model and a popular extension to David-Skene (DS)~\citep{dawid1979maximum} model by making a conditional independence assumption between LFs.
The iBCC model assumes that given the true label $z_i$ of $x_i$, LF labels to $x_i$ are generated independently by different LFs,
\begin{equation}
    p(y_{i1},...,y_{iL}\mid z_i) = \prod_{j=1}^L p(y_{ij}\mid z_i).
    \label{eq:ibcc}
\end{equation}
This was referred as the LF's conditional independence assumption.
However, the underlying independence assumptions prevent the model from capturing correlations between labels from different LF.

\paragraph{Enhanced BCC.}
The EBCC model~\citep{li2019exploiting} is an extension of iBCC, which import $M$ subtypes to capture the correlation between LFs and aggregated the captured correlation by tensor rank decomposition.

The joint distribution of observing the outputs of multiple LFs can be approximated by a linear combination of more rank-1 tensors, known also as tensor rank decomposition~(\cite{hitchcock1927expression}), i.e.,
\begin{equation}
    p(y_{1},...,y_{L}\mid z=k)\approx\sum_{m=1}^M \vec{\pi}_{km} \vec{v}_{1km}\otimes\cdots\otimes \vec{v}_{Lkm},
    \label{eq:tensor_decom}
\end{equation}
where $\otimes$ is the tensor product.
EBCC interpreted the tensor decomposition as a mixture model, where $\vec{v}_{1km}\otimes\cdots\otimes \vec{v}_{Lkm}$ are mixture component shared by all the data samples, and $\pi_{km}$ is the mixture coefficient. 
This comes out that
\begin{equation*}
    p(y_1,...,y_L\mid z) = \sum_{m=1}^M p(g=m\mid z)\prod_{j=1}^L p(y_j\mid z, g=m)
\end{equation*}
here $g$ is an auxiliary latent variable used for indexing mixture components. 
All the mixture components are the result of categorical distribution governed by parameter $\beta_k$ where $\beta_{kk}=a$ and $\beta_{kk'}=b$, which is equivalent to assuming that every LF has correctly labelled $a$ items under every class, and has to make all kinds of mistakes $b$ times.
The $M$ components under class $k$ can be seen as $M$ subtypes, each of which can be used to explain the correlation between LF labels given class $k$~\citep{li2019exploiting}.

\subsection{Multi-class Gaussian Process Classification}
The multi-class Gaussian process (GP) classification model consists of a latent GP prior for each class $\vec{f}=(f_{i1},...,f_{iK})$, where $f_i\sim\GP(m, \Sigma)$, $m$ is the mean over samples, $\Sigma$ is the kernel function.
The conditional distribution is modeled by a categorical likelihood,
\begin{equation}
    p(y_i=k\mid x_i, \vec{f}_i) = h^{(k)}(\vec{f}_i(x_i)),
\end{equation}
where $h^{(k)}(\cdot)$ is a function that maps the real vector of the $\GP$ values to a probability vector.
For $h(\cdot)$, the most common way to form a categorical likelihood is through the softmax transformation
\begin{equation}
    p(y_i=k\mid\vec{f}_i)=\frac{\exp(f_{ik})}{\sum_{k=1}^K\exp(f_{ik})}
\end{equation}
where $f_{ik}$ denotes the $f^k(x_i)$ and for clarity, we omit the conditioning on $x_i$.

\section{METHODS}
In this section, we introduce the proposed \mname model. In a nutshell, it connects the mixture coefficients of the EBCC model with the instance features via a predictive Gaussian process (GP). Then, we introduce a bunch of auxiliary variables to handle the non-conjugation in the model to ensure efficient variational inference. Finally, we present the generative process, joint distribution, and the inference process of the \mname model.

\subsection{Leveraging instance features via mixture coefficient}
In this work, we aim to explicitly incorporate instance features into a statistical label model built upon EBCC.
To attack this problem, we leverage the Gaussian process (GP) classification. 
Specifically, we model the mixture coefficients of EBCC as the output of a GP classifier, which inputs the instance features.
We generate $N\times K\times M$ functions for each data, class, and subtypes, and take the logistic-softmax distribution for each subtype and class to acquire the mixture coefficient for each data. 
In particular, we rewrite the Equation~\ref{eq:tensor_decom} as
\begin{align}
    &p(y_{i1},...,y_{iL}\mid z_i=k)\notag \\ 
    =& \sum_{m=1}^M \pi_{ikm}\left[\vec{v}_{1km}\otimes\cdots\otimes \vec{v}_{Lkm}\right] \notag \\
    \approx&\sum_{m=1}^M h_{\text{softmax}}^{(k,m)}(\sigma(\vec{f}_i)) \left[\vec{v}_{1km}\otimes\cdots\otimes \vec{v}_{Lkm}\right]\notag,
    \label{eq:gp-ebcc}
\end{align}
where $\sigma(\cdot)$ is sigmoid function and $\pi_{ikm} = p(g_i=m\mid z_i)$.
$f_i$ is GP's latent functions for sample $x_i$ with $f_{i} = f(\vec{x}_i)$ and $f_{i}\sim\GP(m_{i}, \Sigma)$. 
We will soon discuss the details and advantages of our usage of GP classifier in the sequel.

\subsection{Handling the Non-conjugate Prior}

Given the proposed model, we would like to infer the true labels via the standard mean field variational inference process following prior work~\citep{li2019exploiting}. 
However, a key challenge that prevents us from performing variational inference is that as a categorical likelihood function, softmax is non-conjugate to the Gaussian prior, so the variational posterior $q(f_{ikm})$ cannot be derived analytically. 
Inspired by \cite{polson2013bayesian, galy2020multi}, we propose to solve the non-conjugate mapping function in the complete data likelihood by introducing a number of auxiliary latent variables such that the augmented complete data likelihood falls into the exponential family, which is conjugate to the Gaussian prior. 

In the following section we (1) decouple the GP latent variables $f_{ikm}$ in the denominator by introducing of a set of auxiliary $\lambda$-variables and the logistic-softmax function, (2) simplify the model likelihood by introducing Poisson random variables, and (3) use a Pólya-Gamma representation of the sigmoid function to achieve the desired conjugate representation of our model.

\paragraph{Decouple GP latent variables.}
Following \citet{galy2020multi}, we first replace the softmax likelihood with the logistic-softmax likelihood,
\begin{equation}
    \pi_{ikm}=h_{\text{softmax}}^{(k,m)}(\sigma(\vec{f}_i)) =\frac{\sigma(f_{ikm})}{\sum_{j=1}^K\sum_{n=1}^M\sigma(f_{ijn})},
    \label{eq:likelihood}
\end{equation}
where $\sigma(z)=(1+\exp(-z))^{-1}$ is the logistic function.
To remedy the intractable normalizer term $\sum_{j=1}^K\sum_{n=1}^M\sigma(f_{ijn})$, we use the integral identity $\frac{1}{x}=\int_0^\infty e^{-\lambda x}d\lambda$ and express the likelihood (\ref{eq:likelihood}) as
\begin{align}
    &h_{\text{softmax}}^{(k,m)}(\sigma(\vec{f}_i)) \notag \\
    =&\sigma(f_{ikm})\int_0^\infty \exp\left(-\lambda_i \sum_{j=1}^K\sum_{n=1}^M \sigma(f_{ijn})\right)d\lambda_i.
\end{align}

By interpreting $\lambda_i$ as an additional latent variable, we obtain the augmented likelihood
{\small
\begin{align}
    p(\pi_{ikm}\mid f_{ikm}, \lambda_i)=\sigma(f_{ikm})\prod_{j=1}^K\prod_{n=1}^M \exp(-\lambda_i\sigma(f_{ijn})), 
    \label{eq:incrop_lambda}
\end{align}
}
here we impose the improper prior $p(\lambda_i)\propto\mathds{1}_{[0,\infty]}, \forall i\in[1,N]$. 
The improper prior is not problematic since it leads to a proper complete conditional distribution, as we will see at the end of the section.

\paragraph{Poisson augmentation}
By leveraging the moment generation function of the Poisson distribution $\Po(\lambda)$
\begin{equation*}
    \exp(\lambda(z-1)) = \sum_{n=0}^{\infty}z^n\Po(z\mid\lambda).
\end{equation*}
Using $z=\sigma(-f)$, we rewrite the exponential factors as,
\begin{align*}
    \exp(\lambda_i \sigma(f_{ikm})) =& \exp(\lambda_i(\sigma(f_{ikm}) - 1)) \\
    =& \sum_{j=1}^K\sum_{n=1}^M(\sigma(-f_{ijn}))^{\upsilon_{ijn}}\Po(\upsilon_{ijn}\mid\lambda_i),
\end{align*}
which leads to the augmented likelihood
\begin{align}
    p(\pi_{ikm}&\mid f_{ikm}, \upsilon_{ikm}, \lambda_i) \notag\\
    &=\sigma(f_{ikm})\cdot\prod_{j=1}^K\prod_{n=1}^M(\sigma(-f_{ijn}))^{\upsilon_{ijn}},
    \label{eq:poisson argum}
\end{align}
where $\upsilon_{ikm}\sim \Po(\lambda_i)$.

\paragraph{Complete with Pólya-Gamma}
In the last step, we aim for a Gaussian representation of the sigmoid function.
The Pólya-Gamma representation allows us for rewriting the sigmoid function as a scale mixture of Gaussian,
\begin{equation}
    \sigma(z)^n = \int_0^\infty 2^{-n}\exp\left(\frac{\upsilon z}{2}-\frac{z^2}{2}\omega\right)\PG(\omega\mid \upsilon,0)
\end{equation}
where $\PG(\omega\mid\upsilon, b)$ is a Pólya-Gamma distribution.
By applying this augmentation to Equation~\ref{eq:poisson argum} we obtain
\begin{align}
    p(\pi_{ikm}&\mid f_{ikm}, \upsilon_{ikm}, \omega_{ikm})\notag \\
    &=\frac{2^{-(\pi_{ikm}+\upsilon_{ikm})}\exp\left\{ \frac{(\pi_{ikm}-\upsilon_{ikm}) f_{ikm}}{2}\right\}}{\exp\left\{\frac{(f_{ikm})^2}{2}\omega_{ikm}\right\}},
\end{align}
where $\omega_{ikm}\sim\PG(\omega_{ikm}\mid \upsilon_{ikm},0)$ are Pólya-Gamma variables.

Finally, the complete conditions of the GPs' $f_{ikm}$ are
\begin{align}
    &p(f_{ikm}\mid \pi_{ikm}, \omega_{ikm}, \upsilon_{ikm}) \notag\\ =&\mathcal{N}\left(f_{ikm}\mid\frac{1}{2}\hat{\Sigma}_{km}\left(\E[\pi_{ikm}]-\E[\upsilon_{ikm}]\right), \hat{\Sigma}_{km}\right),
\end{align}
where $\hat{\Sigma}_{km} = ({\Sigma}_{km}^{-1}+\diag(\E[\omega_{ikm}]))^{-1}$.
For the conditional distribution, $\lambda_i$ we have

\begin{equation}
    p(\lambda_i\mid \vec{\upsilon}_i) = \ga\left(1+\sum_{j=1}^K\sum_{n=1}^M \gamma_{ijn} + 1, K\right),
\end{equation}

where $\ga(\cdot| a, b)$ indicated a gamma distribution with parameter $a$ and $b$. $\gamma_{ijn}$ is the parameter of the joint distribution of $p(\omega_{ikm}, \upsilon_{ikm})$, detailed in Appendix~\ref{apdx:omega and upsilon}.

In summary, by integrating three auxiliary random variable $\lambda_i, \upsilon_{ikm}, \omega_{ikm}$, we successfully turn the posterior of $f_{ikm}$ from non-conjugate softmax to exponential family form, the Gaussian distribution, which is easy to infer by adopting variational inference with $q(f_{ikm})\sim \N(\hat{m}, \hat{\Sigma})$.

\subsection{The Generative Process and Joint Distribution}
\begin{figure}
    \centering
    \includegraphics[width=0.48\textwidth]{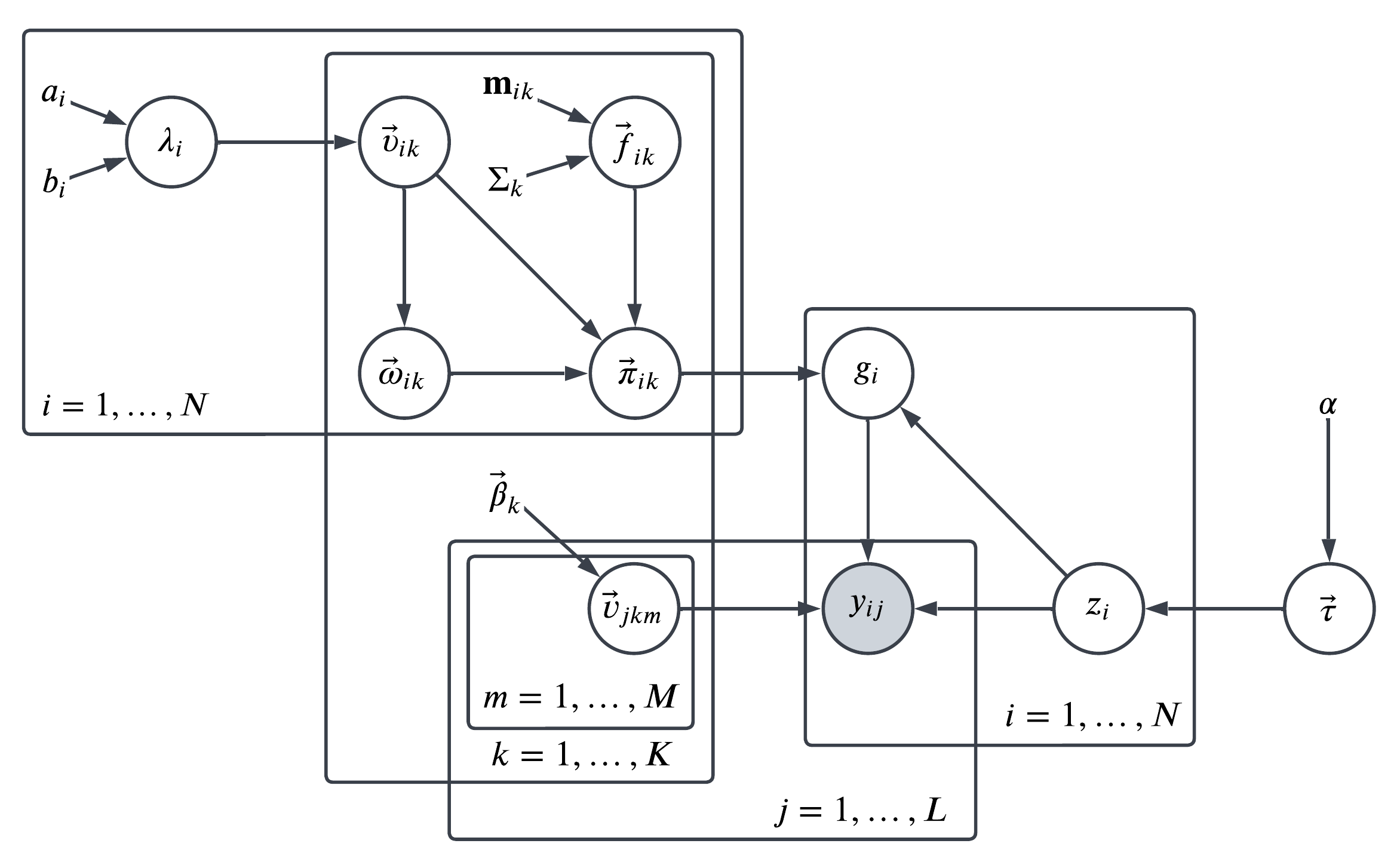}
    \caption{The probabilistic graphical model of \mname.}
    \label{fig:gpebcc}
\end{figure}
Here, we summarize the generative process of the proposed model. We use the GP latent functions $F$ and the corresponding auxiliary variables $\Omega$ and $\Upsilon$ to generate the mixture coefficient $\Pi$.
There are $K\times M$ subtypes in total, and we assume the item $i$ belongs to the $g_i$-th subtype of the class, $z_i$ as in EBCC. The proposed model is shown in Figure~\ref{fig:gpebcc} and its generative process is:

\begin{algorithmic}
\\
\State 1. for $i$ in $1...N$
\State \quad\quad $\lambda_i\sim\ga(a_i, b_i)$
\State \quad\quad for $k$ in $1...K$
\State \quad\quad\quad $\vec{\upsilon}_{ik}\sim\Po(\lambda_i)$
\State \quad\quad\quad $\vec{\omega}_{ik}\sim\PG(\vec{\upsilon}_{ik}, 0)$
\State \quad\quad\quad $\vec{f}_{ik}\sim\mathcal{N}(\vec{m}_{ik}, \Sigma_{k})$
\State \quad\quad\quad $\vec{\pi}_{ik}\sim\ga(\vec{\phi}_{ik}, \vec{\xi}_{ik})$ \\
\State 2. for $k$ in $1...K$
\State \quad\quad for $m$ in $1...M$, for $j$ in $1...L$
\State \quad\quad\quad $\vec{v}_{jkm}\sim\dir(\vec{\beta}_k)$ \\
\State 3. $\vec{\tau}\sim\dir(\vec{\alpha})$ \\
\State 4. for $i$ in $1...N$
\State \quad\quad $z_i\sim\cat(\vec{\tau})$
\State \quad\quad $g_i\sim\cat(\vec{\pi}_{z_i})$
\State \quad\quad for $j\in\mathcal{L}_i$
\State \quad\quad\quad $y_{ij}\sim\cat(\vec{v}_{jz_ig_i})$
\end{algorithmic}
Following the generative process, the joint distribution is 
\begin{align}
\label{eq:gpebcc}
    &p({\bm\lambda}, \Omega, \Upsilon, F, \Pi, V, G, Z, T, Y)\notag \\
    = &p({\bm\lambda})p(\Upsilon\mid{\bm \lambda})p(\Omega\mid\Upsilon)p(F)p(\Pi\mid\Upsilon, \Omega, F)\cdot \notag\\
    &p(V)p(T)p(Z\mid T)p(G\mid\Pi, Z)p(Y\mid Z,G,V).
\end{align}

\subsection{The Inference Algorithm}
The goal of the inference is to find the most likely $Z$ (true labels) given the LF labels $Y$, data features $X$ and all hyperparameters,  
\begin{equation*}
    \argmax_Z p(Z\mid Y, \alpha, a_i, b_i, \vec{m}_{ik}, \Sigma_{km}, \vec{\beta}_k), 
\end{equation*}
which is intractable to solve directly.
Therefore, we adopt a mean-field variational approach that seeks a distribution $q$ that approximates $p({\bm\lambda}, \Omega, \Upsilon, F, \Pi, V, G, Z, T\mid Y, \alpha, a_i, b_i, \vec{m}_{ik}, \Sigma_{km}, \vec{\beta}_k)$, where $q$ is assumed to be factorized as
{\scriptsize
\begin{align*}
    &q({\bm\lambda}, \Omega, \Upsilon, F, \Pi, V, G, Z, T) \\
    =&q({\bm\lambda})q(\Omega, \Upsilon)q(F)q(\Pi)q(V)q(G, Z)q(T) \\
    =&\prod_i\ga(\lambda_i\mid a_i,b_i)\cdot\prod_k\PG(\vec{\omega}_{ik}\mid\vec{\upsilon}_{ik}, \vec{c}_{ik})\Po(\vec{\upsilon}_{ik}\mid\vec{\gamma}_{ik}) \\
    &\mathcal{N}(\hat{m}_{ik}, \hat{\Sigma}_{ik})\ga(\vec{\pi}_{ik}\mid\vec{\phi}_{ik},\vec{\xi}_{ik})\cdot
    \prod_k\prod_m\prod_j\dir(\vec{v}_{kmj}\mid\vec{\mu}_{kmj})\cdot \\
    &\prod_i q(g_i, z_i)\cdot \dir(\vec{\tau}\mid\vec{\nu}).
\end{align*}}

Since the joint distribution is fully factorized in $q$, it is easy to solve $\argmax_Z q(Z)$ by finding $k$ that maximizes every individual $q(z_i=k)$, i.e. $\Tilde{z}_i=\argmax_k q(z_i=k)$.

Let $\rho_{ikm}=q(z_i=k, g_i=m)$, then follow the standard mean-field variational Bayes steps, we can derive the update rules shown below
\begin{align*}
    \rho_{ikm} &= e^{\E_q [\log\tau_k] +\E_q[\log\pi_{ikm}]+\sum_{j\in\mathcal{L}_i}\E_q[\log v_{kmjy_{ij}}]} \\
    q(z_i=k) &= \sum_m\rho_{ikm} \\
    \mu_{kmj} &= \beta_{kl} + \sum_{i\in\mathcal{N}_j}\rho_{ikm}\mathds{1}[y_{ij}=l]\\
    \nu_k &= \alpha_k + \sum_i q(z_i=k) \\
    \phi_{ikm} &= \rho_{ikm}+1 \\
    \xi_{ikm} &= \log2 - \frac{\hat{m}_{ikm}}{2} \\
    \hat{\Sigma}_{km} &= (\Sigma_{km}^{-1}+\diag(\E[\omega_{ikm}]))^{-1} \\
    \hat{m}_{ikm} &= \frac{1}{2}\hat{\Sigma}_{km}(\phi_{ikm}/\xi_{ikm}-\E[\upsilon_{ikm}]) \\
    c_{ikm} &= \sqrt{\hat{m}_{ikm}^2 + \hat{\Sigma}_{km}(i,i)} \\
    \gamma_{ikm} &= \frac{\exp(\phi(a_i)-\frac{\hat{m}_{ikm}}{2})}{\beta_i\cosh(\frac{c_{ikm}}{2})} \\
    a_i &= \sum_k\sum_m\gamma_{ikm} + 1 \\
    b_i &= K
\end{align*}
The expectations are calculated as follows
\begin{align*}
    &\E_q[\log\tau_k] = \psi(\nu_k)-\psi(\sum_k\nu_k)\\
    &\E_q[\log\pi_{ikm}] = \psi(\phi_{ikm}) - \log(\xi_{ikm}) \\
    &\E_q[\log v_{jkml}] = \psi(\mu_{jkml}) - \psi(\sum_l\mu_{jkml}).
\end{align*}
While updating $\hat{\Sigma}_{ikm}$ through variational inference, one has to calculate the inverse of $(\Sigma_{km}^{-1} + \diag(\E[\omega_{ikm}]))$, a $N\times N$ matrix, $K\times M$ times every inference step, which could be prohibitively slow for large-scale dataset.
To address this issue, we adopt the Lanczos algorithm~\citep{golub2013matrix} to acquire a low-rank approximation of $(\Sigma_{km}^{-1} + \diag(\E[\omega_{ikm}]))^{-1}$ and achieve at least 10 times acceleration.

Briefly, the Lanczos algorithm factories a symmetric matrix $A\in\mathbb{R}^{n\times n}$ as $QTQ^\top$, where $T\in\mathbb{R}^{n\times n}$ is symmetric tridiagonal and $Q\in\mathbb{R}^{n\times n}$ is orthonormal by using a probe vector $\textbf{b}$ and computes an orthogonal basis of the Krylov subspace $\mathcal{K}(A, \textbf{b})$
\begin{equation*}
    \mathcal{K}(A, \textbf{b})=\text{span}\{\textbf{b}, A\textbf{b}, A^2\textbf{b},...,A^{n-1}\textbf{\textbf{b}}\}
\end{equation*}
Applying Gram-Schmidt orthogonalization to these vectors produces the columns of $Q, [\textbf{b}/\|\textbf{b}\|, \textbf{q}_2, \textbf{q}_3,...,\textbf{q}_n]$ (here $\|\textbf{b}\|$ is the Euclidean norm of $\textbf{b}$). The orthogonalization coefficients are collected into $T$. Because $A$ is symmetric, each vector needs only be orthogonalized against the two preceding vectors, which results in the tridigonal structure of $T$~\citep{golub2013matrix}. The orthogonalized vectors and coefficients are computed in an iterative manner. $k$ iterations produce the first $k$ orthogonal vectors of $Q_k=[\textbf{q}_1,...,\textbf{q}_k]\in\mathbb{R}^{n\times k}$ and their corresponding coefficients $T_k\in\mathbb{R}^{k\times k}$. These $k$ iterations require only $O(k)$ matrix vector multiplies with the original matrix A.
\section{EXPERIMENT}

\subsection{Implementation Details}
\paragraph{Initialization.}
\mname has 5 parameters, $q(z_i=k), \vec{\rho}_{ik}, \hat{\Sigma}_{km}, \hat{m}_{ikm}$, and $a_i$ to be initialized.
For $q(z_i=k)$, our initialization is similar to EBCC: we first initialize $q(z_i=k)$ by majority voting, i.e. $q(z_i = k) = \frac{1}{|L_i|}\sum_{j\in L_i} 1[y_{ij}=k]$, then multiply it with a random vector drawn from $\dir(1_M)$ to initialize $\vec{\rho}_{ik}$.
We calculate the pair-wise cosine similarity of the input features to initialize $\hat{\Sigma}_{km}$.
Finally, we initialize $\hat{m}_{ikm}, a_i$ with an uninformative prior $\text{Uniform}(0,1)$.

\paragraph{Hyperparameter settings.}
\mname has 5 hyperparameters, $\beta_{kk}, \beta_{kk'}$ for initializing $\vec{\mu}_{jkm}$, $\vec{\alpha}$ for initializing $\vec{\tau}$ and the number of subtype $M$.
We set $\beta_{kk}=N\times M \times C$, $\beta_{kk'}=1$, $k\neq k'$, where $C$ is the number of correct labels that LFs gave in each subtype and class. We set $C=1000$ to encode that we believe LFs are better than random guessing.
Following EBCC, we set $\alpha_k=\sum_i q(z_i=k)^{(0)}$ where $q(z_i=k)^{(0)}$ is the MV initialization for $q(z_i=k)$ because MV can provide a reliable estimate of the class portion in the dataset. 
For the number of subtypes, we set $M = 3$. 
The key reason of we give a small number of subtypes is that subtypes are learned to capture correlation patterns and a large $M$ increase the number of parameters, increasing the risk of overfitting.

\subsection{Compared Methods}
We compare our method \mname against existing label models implemented in the WRENCH benchmark~\citep{zhang2021wrench} as well as iBCC and EBCC. 
For all the baselines, we use the default parameter without hyperparameter tuning because we do not assume a gold validation set.
We list the involved baselines as follows:
\begin{itemize}
    \item \textbf{Majority Voting (MV)}. The predicted label of each data point is the most common label given by LFs.
    
    \item \textbf{Data Programming (DP)}~\citep{ratner2016data}. DP models the distribution $p(Y, Z)$ as a factor graph. It can describe the distribution in terms of pre-defined factor functions, which reflects the dependency of any subset of random variables. The log-likelihood is optimized by SGD where the gradient is estimated by Gibbs sampling, similarly to contrastive divergence~\citep{salakhutdinov2010efficient}.
    
    \item \textbf{MeTaL}~\citep{ratner2019training}. MeTal models the distribution via a Markov Network and recover the parameters via a matrix completion-style approach. The latest version of the popular Snorkel system\footnote{\url{https://github.com/snorkel-team/snorkel}} adopts MeTaL as its default label aggregation method.
    
    \item \textbf{FlyingSquid (FS)}~\citep{fu2020fast}. FS models the distribution as a binary Ising model, and a Triplet Method is used to recover the parameters. Notably, FS is designed for binary classification and the author suggested applying a one-versus-all reduction repeatedly to apply the core algorithm. 
    
    \item \textbf{Dawid and Skene's model (DS)}~\citep{dawid1979maximum}. DS models the confusion matrix of each worker regarding the ground truth labels. This method is widely used in crowdsourcing and is the recommended method for classification tasks in a benchmark on crowdsourcing~\citep{zheng2017truth}.
    
    \item \textbf{Independent BCC (iBCC)}~\citep{kim2012ibcc}. iBCC models the relation between workers' annotation and the ground true label by worker independent assumption, and the relation can be solved by Gibbs sampling, mean-filed variational Bayes and expectation propagation.
    
    \item \textbf{Enhanced BCC (EBCC)}~\citep{li2019exploiting}. Based on iBCC, EBCC models the latent correlation between workers by adding subtypes that have significant potential to improve truth inference.
    
    \item \textbf{Denoise}~\citep{ren2020denoising}. \textit{Denoise} adopts an attention network to aggregate over weak labels, and use a neural classifier to leverage the data features.
    These two components are jointly trained in an end-to-end manner.
    
    \item \textbf{WeaSEL}~\citep{ruhling2021end}. WeaSEL shares similar model architecture as Denoise with a new objective to optimize the two components jointly.
\end{itemize}

\subsection{Synthetic Dataset}

In this section, we use synthetic datasets to show that leveraging instance features makes a statistical label model robust to the dataset size, and to answer the question of when is leveraging instance features helpful in improving the performance of label aggregation?

\paragraph{Leveraging instance features makes a statistical label model robust to the dataset size.}
\begin{figure}
\centering
\hspace{-8mm}
\includegraphics[width=0.3\textwidth]{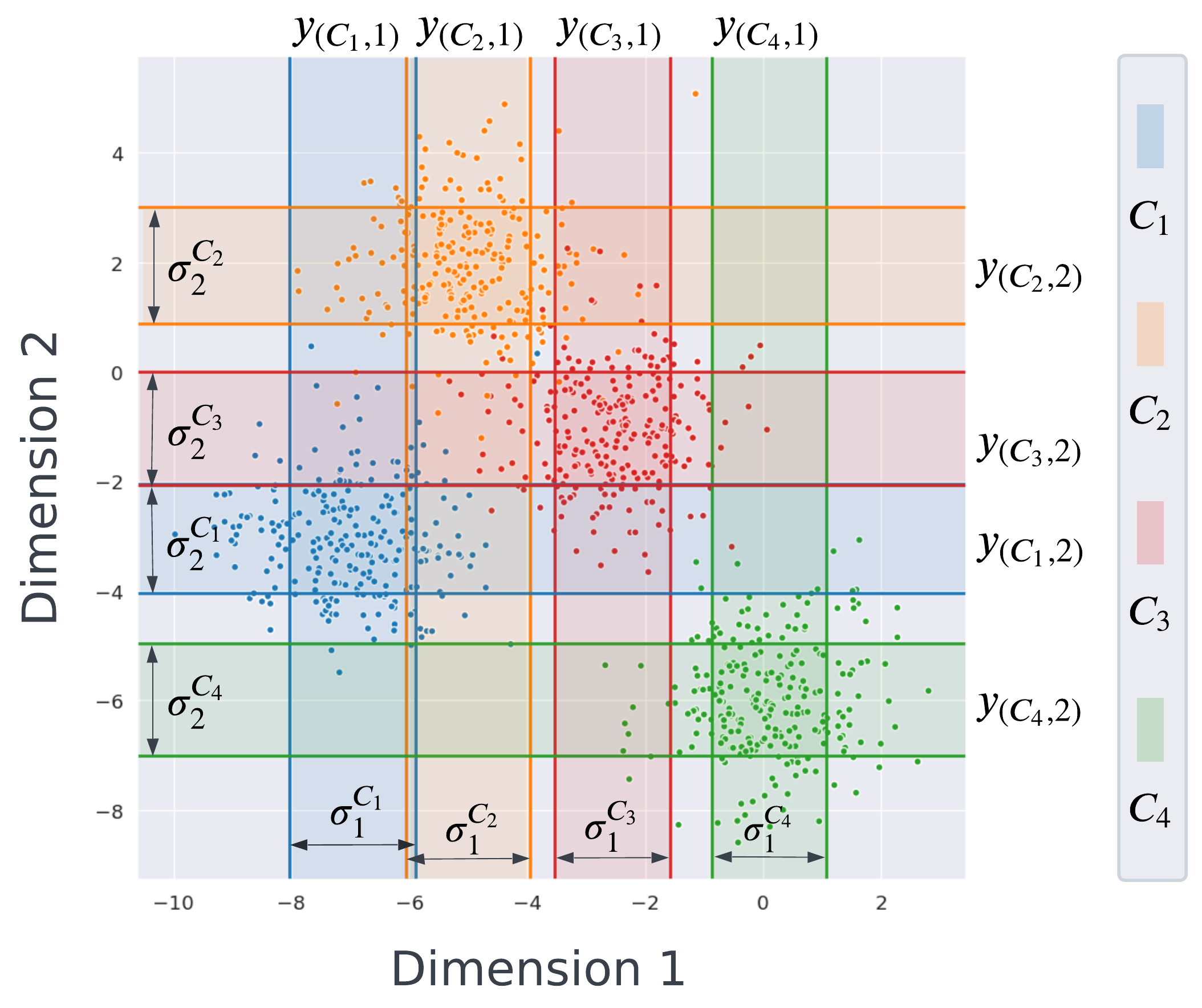}
\caption{An example of decision space of each LF in the synthetic dataset.}
\label{fig:covered_syn}
\end{figure}
The authors of EBCC showed that the performance of EBCC drops dramatically when the size of synthetic dataset increases, and they hypothesized that it is an optimization problem and EBCC gets stuck on bad local optima~\citep{li2019exploiting}. We argue that a statistical label model leveraging instance features like \mname does might suffer less from bad local optima because it introduces a strong yet realistic inductive bias: similar data tend to have similar correlation pattern, which serves as an implicit constraint and potentially avoid bad local optima.

To confirm our hypothesis, we generated synthetic datasets consisting of 4 classes, \ie, $\{C_1, C_2, C_3, C_4\}$, with different dataset sizes, \ie, $\{1, 5, 10, 15, 20\}\times10^3$, for evaluation. Data feature $\vec{x}_i$ in each class are sampled from 4 different Gaussian distributions with 2 independent features $x_{i1}, x_{i2}$, and $\mathcal{X}^{C_j}$ denotes the set of synthetic data belonging to the class $C_j$.
Then, we generate 8 \emph{unipolar} LFs, $\vec{y}=[y_{(C_1,1)}, y_{(C_1,2)}, ..., y_{(C_4,1)}, y_{(C_4,2)}]$ with 2 for each class. 
Specifically, a LF $y_{(C_j, k)}$ assign label $C_j$ or abstention ($-1$) to an individual data point based on the $k$-th dimension of the data feature.
For a data point $x_i$, the output of a LF $y_{(C_j, k)}$ is: 
\begin{equation*}
    y_{(C_j, k)}=\left\{
    \begin{aligned}
    C_j, & \ \text{if} \ \ \mu^{C_j}_{k} - \sigma_k^{C_j} <x_{ik}< \mu^{C_j}_{k} + \sigma_k^{C_j}\\
    -1, & \ \text{otherwise}
    \end{aligned},
    \right.
\end{equation*}
where $-1$ means the LF abstaining from voting, and $\mu^{C_j}$ and $\sigma^{C_j}$ indicate the mean and standard deviation respectively of generated data points $x\in\mathcal{X}^{C_j}$. We use subscript $k$ to indicate the value of the $k$-th dimension value of $\mu^{C_j}$ or $\sigma^{C_j}$.
We provide an example of the generated synthetic data and LFs as in Figure.~\ref{fig:covered_syn}.

\begin{figure}
    \centering
    \includegraphics[width=0.45\textwidth]{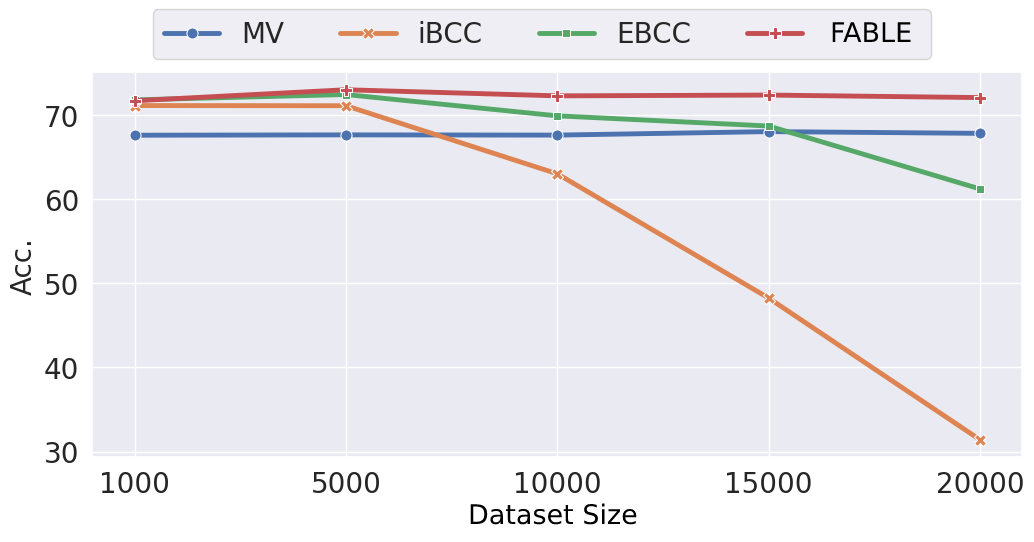}
    \caption{Performance comparison on synthetic dataset, the results are average over 100 runs on each dataset.}
    \label{fig:perf_syn}
\end{figure}

\begin{figure}[t!]
    \centering
    \hspace{-8mm}
    \includegraphics[width=0.3\textwidth]{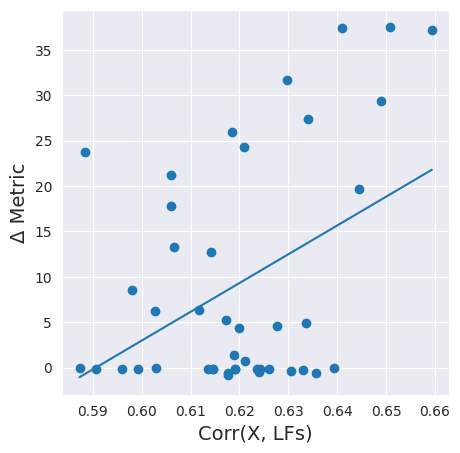}
    \caption{Relation between feature-LF correlation and performance gap between EBCC and \mname on synthetic dataset.}
    \label{fig:rela_syn}
\end{figure}
\begin{table*}[h!]
\centering
\caption{Dataset statistics}
\setlength{\tabcolsep}{2pt}
\renewcommand{\arraystretch}{1.1}
\begin{tabular*}{\textwidth}{@{\extracolsep{\fill}}c|ccccccccccc}
\toprule\hline
Dataset & IMDB  & Youtube & SMS  & CDR & Yelp    & Commercial & Tennis & TREC & SemEval & ChemProt & AG News \\ \hline\hline
Metric  & F1    & F1      & F1   & F1    & F1    &F1          & F1     & Acc  & Acc     & Acc      & Acc     \\ \hline
\#Class & 2     & 2       & 2    & 2     & 2     & 2          & 2      & 6    & 9       & 10       & 4    \\ \hline
\#LF    & 5     & 10      & 73   & 33    & 8     & 4          & 4      & 68   & 164     & 26       & 9    \\ \hline
\#Data  & 25,000 & 1,956    & 5,571 & 14,023 & 38,000 & 81,105      & 20,256  & 2,965 & 2,641    & 16,075    & 120,000\\ \hline\bottomrule
\end{tabular*}
\label{tab:datasets}
\end{table*}
\begin{table*}[h!]
\centering
\caption{Comparison among different methods on benchmark dataset. \best{Red} and \secbest{blue} indicate the best and the second-best result of each method.}
\renewcommand{\arraystretch}{1.1}
\begin{tabular*}{\textwidth}{@{\extracolsep{\fill}}c | ccccccccc | c}
\toprule\hline
Dataset    & MV              & DS              & DP              & FS             & MeTaL    & iBCC            & EBCC            & WeaSEL      & Denoise    & \mname           \\ \hline\hline
IMDB       & 72.19           & 70.32           & 72.26           & 72.45          & 72.10    & 66.74           & \secbest{74.18} & 67.99       & \best{83.61}      & 73.96 \\ \hline
Youtube    & 80.74           & 81.62           & 74.90           & 78.83          & 76.63    & 73.83           & \secbest{86.57} & 0.00        & 83.79             & \best{88.56} \\ \hline
SMS        & 32.80           & 43.58           & 32.79           & 30.05          & 32.27    & 48.41           & 48.41           & 0.00        & \best{87.40}      & \secbest{48.81} \\ \hline
CDR        & \secbest{63.16} & 53.64           & 53.37           & \best{64.81}   & 43.28    & 9.59            & 23.89           & 0.00        & 61.72      & 62.15 \\ \hline
Yelp       & \secbest{73.42} & 71.93           & 73.13           & \best{73.95}   & 69.97    & 68.59           & 72.87           & 66.67       & 66.67      & 72.50 \\ \hline
Commercial & 84.23           & \best{88.31}    & 76.43           & 80.86          & 78.61    & 76.83           & 76.43           & 0.00        & \secbest{87.66}      & 86.62 \\ \hline
Tennis     & \secbest{83.82} & 83.56           & \best{84.29}    & 83.31          & 83.62    & 83.64           & 83.67           & 77.32        & 19.71      & 83.63  \\ \hline
TREC       & 52.35           & 47.18           & \best{55.07}    & 48.32          & 41.94    & 41.91           & 46.94           & 27.60       & 46.17      & \secbest{53.20}  \\ \hline
SemEval    & \best{78.35}    & 73.53           & 73.53           & 11.20          & 72.69    & 73.53           & 73.53           & 30.19       & 67.12      & \secbest{74.32}  \\ \hline
ChemProt   & 47.96           & 38.82           & 45.71           & 46.25          & \best{49.76} & 31.84       & 33.80           & 31.84       & 45.54      & \secbest{48.35}  \\ \hline
AG News    & 63.85           & \secbest{63.95} & 63.56           & 63.63          & \best{64.15} & 25.00       & 55.94           & 25.00       & 50.49      & 62.74 \\ \hline\hline

Avg. Perf. & \secbest{66.63} & 65.13           & 64.26           & 59.42          & 61.62    & 54.54           & 61.48           & 46.66       & 63.63      &\best{68.55} \\\hline

Avg. Rank  & \secbest{3.36}  & 4.72            & 4.63            & 5.09           & 5.90     & 7.09            & 4.72            & 9.45           & 5.36  &\best{3.09} \\
\hline\bottomrule
\end{tabular*}
\label{tab:mainres}
\end{table*}

All results are reported in Figure.~\ref{fig:perf_syn}, solid lines show the averaged accuracy of 100 runs. We compare \mname against MV, IBCC, and EBCC to show the advantages of leveraging instance features.
From the results, we can see that the performance of MV is quite stable because the MV method does not involve any learnable parameter.
In addition, the performance of previous BCC models (iBCC, EBCC) drop dramatically when the dataset size increases, and the EBCC decreases much slower than iBCC, which is aligned with the findings of EBCC paper.
Finally, our proposed method \mname is consistently better than MV and more importantly, it is as stable as MV, which confirms our hypothesis that the regulation effect of leveraging instance features could avoid bad local optima compared to other BCC models without instance features.

\paragraph{When are instance features useful?}
\label{sec:lf_corr}
We are curious about when the instance features are useful and could lead to better performance of label aggregation. We hypothesize that the superiority of leveraging instance features is positively correlated to the correlation between instance features and the correctness of LFs. We formally define such a correlation as follows:
\begin{equation}
    \text{Corr}(X, \text{LFs}) = \frac{1}{L}\sum_{j=1}^L \text{dCor}(X_j, R_j),
\end{equation}
where $X_j\in\{\vec{x_i}\mid y_{ij}\neq -1, i\in [N]\}$, $R_j=\{\mathds{1}[y_{ij}=z_i]\mid y_{ij}\neq -1, i\in [N]\}$, and $\text{dCor}(\cdot, \cdot)$ is the distance correlation. Such a correlation could reflect the level of instance features being indicative of the correlation patterns a BCC model tend to capture because the correlation pattern is indeed an instantiation of confusion matrix of true label $z$ and LF label $y$ and the correctness of LF, \ie, $\mathds{1}[y_{ij}=z_i]$ can be treated as a simplified version of such a confusion matrix.

To verify the above claim, we fix the dataset size to be $1,000$ and modify the generative process of synthetic dataset by introducing a hyperparameter $\psi$ to generate different LFs.
Specifically, each $y_{(C_j, k)}$ now annotates data according to 
\begin{equation*}
    y_{(C_j, k)}=\left\{
    \begin{aligned}
    C_j, & \ \text{if} \ \ \mu^{C_j}_{k} - \psi\sigma_k^{C_j} <x_{ik}< \mu^{C_j}_{k} + \psi\sigma_k^{C_j}\\
    -1, & \ \text{otherwise}
    \end{aligned}.
    \right.
\end{equation*}
To generate a synthetic dataset, we randomly sample eight $\psi\sim\text{Uniform}(1,3)$, one for each LF. We generate 50 synthetic datasets in this way and calculate the performance gain of \mname over EBCC as $\Delta$ Metric because we would like to see how the correlation $\text{Corr}(X, \text{LFs})$ affects the superiority of leveraging instance features (Note that the \mname can be seen as EBCC with instance features incorporated).

The result in Figure~\ref{fig:rela_syn} shows that the performance gap between \mname and EBCC is positively related to the correlation $\text{Corr}(X, \text{LFs})$.
The inset line is generated by linear regression over all recorded results.
To further prove our hypothesis, we calculate the Pearson's correlation coefficient $r$ between $\text{Corr}(X, \text{LFs})$ and $\Delta$ Metric.
The $r=0.469$ with $p$-value $< 0.01$, which implies that there exists a positive relationship between the performance gain of leveraging instance features and the correlation $\text{Corr}(X, \text{LFs})$ with high confidence.
This finding provides practitioners with some insights of when to incorporate instance features in a statistical model: that is, if the correlation between LFs and true label (in other words, the confusion matrix) is highly dependent on the instance feature, then it is beneficial to incorporate instance features as \mname does.

\subsection{Benchmark Datasets}
We conduct experiments on eleven classification datasets across diverse domains (\eg, income/sentiment/span/relation/question/topic classification tasks) from the WRENCH benchmark~\citep{zhang2021wrench}.
The WRENCH benchmark splits each dataset into training/validation/test sets, while we follow \cite{li2019exploiting} to adopt a transductive setting, \ie, we perform model learning and evaluation on the whole dataset without any ground truth label.
The detail of all benchmark datasets are listed in Table.~\ref{tab:datasets}.
In the case of text dataset, we use RoBERTa~\citep{liu2019roberta} to extract features following \cite{zhang2021wrench}, while for other datasets, we use the original features coupled with each dataset.

\paragraph{Results.}
We report the performance comparison over 11 datasets as well as each method's averaged score of evaluation metrics and averaged ranking over datasets in Table~\ref{tab:mainres}.
\mname achieves the highest average performance and ranking, outperforming all the baselines.
On most datasets, \mname is either the best or second-best method.
And the majority voting is the second-best method in average due to its simplicity and the heterogeneity of the WRENCH benchmark~\citep{zhang2021wrench}.
We identified two failure cases of EBCC, namely, CDR and ChemProt, where EBCC performs much lower than \mname.
After examining the learned parameters of EBCC on CDR, we found that it learns a highly skewed distribution of $\eta_{km}$, which is used for generating the mixture coefficient for EBCC.
In particular, one of the values of $\eta_{km}$ is much higher than others, which indicates that EBCC failed to capture the multiple distinct correlation patterns between the LFs and true label.
It may be caused by the fact that in CDR, two of the LFs have much higher coverage (portion of non-abstention votes) than the others and such high-coverage LFs might have more complex and instant-dependent correlation patterns than low-coverage ones as they cover more diverse instants, but as an instant-independent model, EBCC may not be able to identify the multi-modal instant-dependent correlation patterns.
In contrast, \mname parameterizes the mixture coefficient using instance features, enabling it to capture the complex instant-dependent patterns.
And for the case of ChemProt, we had a similar observation.
Additionally, We compare \mname with two neural network-based methods: WeaSEL and Denoise, which also leverage instance features.
The results show that neural network-based methods under-perform \mname in most cases.
The key reason could be that although they use sophisticated neural networks to incorporate instance feature and introduce extra parameters, they highly rely on a gold validation set for hyperparameter tuning and early stopping~\citep{zhang2021wrench}, which is unavailable in our setup as we do not assume any gold labeled data.
\section{CONCLUSION}
In this work, we developed a statistical label model for label aggregation in Programmatic Weak Supervision with the goal of leveraging instance features in statistical modeling.
Built upon a recent mixture model called EBCC, our model, \mname (Feature-Aware laBeL modEl), achieves this goal by introducing a predictive Gaussian process to output the mixture coefficient based on instance features.
The efficacy of \mname is demonstrated in extensive experiments on synthetic datasets.
We also showed that the performance gain of \mname over EBCC is positively related to the level of instance features being indicative of correlation patterns between the LFs' votes and the true label.
We compared \mname with 7 baselines on 11 benchmark datasets from various domains, and \mname achieves the best averaged performance.

\paragraph{Social impact.} All the dataset we used are public available and does not involve any human object. We do not foresee any negative social impact of our work. Our study aims to advance the field of Programmatic Weak Supervision, which can reduce human efforts in collecting training labels when developing ML models. Thus, we believe that our work has positive social impact by making the development of ML model easier.

\paragraph{Limitations.} Our proposed method involves a predictive Gaussian process which could be inefficient to scale up for large dataset compared with simple method like majority voting. However, the rich literature of accelerating Gaussian process could inspire future improvement on our method regarding the scalability.

\bibliographystyle{plainnat}
\bibliography{reference}

\appendix
\onecolumn
\aistatstitle{Leveraging Instance Features for Label Aggregation in Programmatic Weak Supervision: \\
Supplementary Materials}

\section{Inference of EBCC}
According to the generative process and graphical model, we have:

\begin{align*}
    p(\pi, V, \tau, Z, G, Y \mid a_\pi, \alpha, \beta)
    &= p(\pi \mid a_\pi)p(V\mid \beta)\cdot p(\tau\mid \alpha)p(Z\mid \tau)p(G\mid\pi, Z)p(Y\mid Z,G,V) \\
    &\propto \prod_{k=1}^K\prod_{m=1}^M\pi_{km}^{a_\pi -1} \cdot \prod_{j=1}^L\prod_{k=1}^K\prod_{m=1}^M\prod_{l=1}^K v_{jkml}^{\beta_{kl}-1} \cdot \\ 
    &\ \ \ \prod_{k=1}^K\tau_k^{\alpha_k-1} \cdot \prod_{i=1}^N\tau_{z_i}\cdot\prod_{i=1}^N\pi_{z_i g_i}\cdot\prod_{i=1}^N\prod_{j\in\mathcal{W}_i} v_{jz_ig_iy_{ij}}.
\end{align*}

To find the most likely $Z$ given the LF labels $Y$ and all hyperparameters, EBCC use fully Bayesian inference algorithm, and adopt a mean-field variational approach to find a distribution $q$ that approximates $p(\pi, V, \tau, Z, G, Y \mid a_\pi, \alpha, \beta)$:
\begin{align*}
\operatorname{argmax}_Z p(Z\mid Y, a_\pi, \alpha, \beta)
&=\operatorname{argmax}_Z\sum_G \int p(Z\mid Y, a_\pi, \alpha, \beta)d\tau d\pi dV \\
&\approx \operatorname{argmax}_Z\sum_G\int q(\tau, Z, G, \pi, V)d\tau d\pi dV \\
&=\operatorname{argmax}_Zq(Z).
\end{align*}

And $q$ can be factorized as 
\begin{align*}
\label{eq:qebcc}
q(\tau, Z, G, \pi, V) 
&= q(\tau)q(Z, G)q(\pi)q(V) \\
&= \operatorname{Dir}(\tau\mid\nu)\cdot \prod_{i=1}^N q(z_i,g_i)\cdot\prod_{k=1}^K\operatorname{Dir}(\pi_k\mid\eta_k)\cdot\prod_{k=1}^K\prod_{m=1}^M\prod_{j=1}^L\operatorname{Dir}(v_{kmj}\mid\mu_{kmj}).
\end{align*}

So, the ELBO is
\begin{align*}
&\E_q\left[\log p(\tau, Z, G, Y, \pi, V\mid a_\pi, \alpha, \beta) - \log q(\tau, Z, G, \pi, V)\right] \\ 
&= \sum_{k=1}^K(\nu_k-1)\E_q[\log\tau_k] + \sum_{k=1}^K\sum_{m=1}^M(\eta_{km}-1)\E_q[\log\pi_{km}] + \sum_{j=1}^L\sum_{k=1}^K\sum_{m=1}^M\sum_{l=1}^K(\mu_{jkml}-1)\E_q[\log v_{jkml}] \\
&-\log B(\alpha)-K\log B(a_\pi\textbf{1}_M)-WM\sum_{k=1}^K\log B(\beta) + H(\operatorname{Dir}(\tau\mid\nu))+\sum_{i=1}^N H(q(z_i, g_i))  \\ 
&+\sum_{k=1}^K H(\operatorname{Dir}(\pi_k\mid\eta_k)) + \sum_{j=1}^L\sum_{k=1}^K\sum_{m=1}^M H(\operatorname{Dir}(v_{jkm}\mid\mu_{jkm}))
\end{align*}

\subsection{Update rule for $q(Z, G)$ and $q(Z)$}
To get the optimal solution for $q(Z, G)$, we need to find the exponential family form of its posterior. According to the PGM, we have
\begin{align*}
    \log q(z_i = k, g_i = m)
    &=\E_{\tau, \pi, V}\left[ p(z_i\mid \tau) \cdot p(g_i\mid\pi_k, z_i) \cdot p(y_{ij} \mid z_i,g_i,v_{jkm}) \right] \\
    &\propto \E_{\tau, \pi, V}\left[ \log(\tau_{k}) + \log(\pi_{km}) + \log(v_{jkm}) \right],
\end{align*}
where the natural parameter $\eta = \E_{\tau, \pi, V}\left[ \log(\tau_{k}) + \log(\pi_{km}) + \log(v_{jkm}) \right]$. According to the inverse parameter mapping for categorical distribution with base measure $h(x)=1$, we have
\begin{align*}
    \rho_{ikm} 
    &= e^\eta \\
    &= \exp\Big\{ \E_\tau [\log\tau_k] + \E_\pi[\log\pi_{km}] + \E_V[\log v_{jkm}] \Big\} \\
    &=\exp\left\{ \Psi(\nu_k)-\Psi\left(\sum_{k=1}^K\nu_k\right) + \Psi(\eta_{km}) - \Psi\left(\sum_{m=1}^M\eta_{km}\right) + \Psi(\mu_{jkml}) - \Psi\left(\sum_{l=1}^K\mu_{jkml}\right) \right\}.
\end{align*}
Note that for $\theta \sim \operatorname{Dir}(\alpha_1, \cdots, \alpha_k)$, we have
\begin{equation*}
    \E[\log(\theta_k)\mid\alpha] = \Psi(\alpha_k) - \Psi\left(\sum_{i=1}^N\alpha_i\right),
\end{equation*}
where $\Psi(\cdot)$ is the digamma function. Since $\rho_{ikm}=q(z_i=k, g_i=m)$, we can also easily get
\begin{equation*}
q(z_i=k) = \sum_{m=1}^M q(z_i=k, g_i=m) = \sum_{m=1}^M\rho_{ikm}
\end{equation*}

\subsection{Update rule for $q(\tau)$}
\label{sec:nu}
To get the optimal solution for $q(\tau)$, we need to find the exponential family form of its posterior. According to the PGM, we have
\begin{align*}
    \log q(\tau_k) 
    &= \E_Z\left[ \log \left(p(\tau_k)\cdot \prod_{i=1}^N p(z_i=k\mid\tau_k) \right) \right] \\
    &=  (\alpha_{k} - 1)\log\tau_k + \sum_{i=1}^N \E_Z\left[q(z_i=k)\right]\log\tau_k ,
\end{align*}
where the natural parameter $\eta=\alpha_k-1 + \sum_{i=1}^N q(z_i=k)$ and $\gamma_{ik} = q(z_i=k)$. According to the inverse parameter mapping for Dirichlet distribution with base measure $h(x)=1$, we have
\begin{align*}
    \nu_k 
    &= 1 + (\alpha_k - 1) + \sum_{i=1}^N \E_Z\left[q(z_i=k)\right] \\
    &= \alpha_k + \sum_{i=1}^N \gamma_{ik}
\end{align*}

\subsection{Update rule for $q(\pi)$}
\label{sec:eta}
To get the optimal solution for $q(\pi)$, we need to find the exponential family form of its posterior. According to the PGM, we have
\begin{align*}
    \log q(\pi_{km}) 
    &= \E_{Z, G}\left[\log\left( p(\pi_{km})\cdot \prod_{i=1}^N p(g_i=m \mid \pi_{km}, z_i=k) \right)\right] \\
    &=  (a_\pi - 1)\log\pi_{km} + \sum_{i=1}^N \E_{Z, G}\left[q(z_i = k, g_i = m)\right]\log\pi_{km} ,
\end{align*}
where the natural parameter $\eta=a_\pi - 1 + \sum_{i=1}^N q(z_i=k, g_i=m)$ and $\rho_{ikm} = q(z_i=k, g_i=m)$. According to the inverse parameter mapping for Dirichlet distribution with base measure $h(x)=1$, we have
\begin{align*}
    \eta_{km} 
    &= 1 + (a_\pi - 1) + \sum_{i=1}^N \E_{Z, G}\left[q(z_i = k, g_i = m)\right] \\
    &= a_\pi + \sum_{i=1}^N \rho_{ikm}
\end{align*}

\subsection{Update rule for $q(v)$}
To get the optimal solution for $q(v)$, we need to find the exponential family form of its posterior. According to the PGM, we have
\begin{align*}
    \log q(v_{jkml}) 
    &= \E_{Z, G}\left[ \log\left( p(v_{jkml})\cdot \prod_{i\in\N_j} p( y_{ij} \mid v_{jkml}, g_i=m, z_i=k) \right) \right] \\
    &= \E_{Z, G}\left[ (\beta_{kl} - 1)\log v_{jkml} + \sum_{i\in\N_j} q(g_i=m, z_i=k)\cdot\delta(y_{ij}, 1)\log v_{jkml} \right] \\
\end{align*}
where the natural parameter $\eta=(\beta_{kl} - 1) + \sum_{i\in\N_j} q(g_i=m, z_i=k)\cdot\delta(y_{ij}, 1)$ and $\rho_{ikm} = q(z_i=k, g_i=m)$. According to the inverse parameter mapping for Dirichlet distribution with base measure $h(x)=1$, we have
\begin{align*}
    \mu_{jkml} 
    &= 1 + (\beta_{kl} - 1) + \sum_{i\in\N_j} q(g_i=m, z_i=k)\cdot\delta(y_{ij}, 1) \\
    &= \beta_{kl} + \sum_{i\in\N_j} \rho_{ikm}\cdot\delta(y_{ij}, 1)
\end{align*}

\section{Inference of \mname}

We integrate Gaussian process into the EBCC as \mname with 3 auxiliary variables: $\lambda, \Upsilon, \Omega$. According to the PGM in Fig.~\ref{fig:gpebcc}, we can decompose the prior as

\begin{align}
\label{eq:gpebcc}
    p({\bm\lambda}, \Omega, \Upsilon, F, \Pi, V, G, Z, T, Y)\notag
    &= p({\bm\lambda})p(\Upsilon\mid{\bm \lambda})p(\Omega\mid\Upsilon)p(F)p(\Pi\mid\Upsilon, \Omega, F)\cdot \notag\\
    &\ \ \ \ p(V)p(\tau)p(Z\mid \tau)p(G\mid\Pi, Z)p(Y\mid Z,G,V)
\end{align}

According to Eq.~\ref{eq:gpebcc}, which is based on EBCC, we only need to change the inference process of the variational distribution of mixture coefficient $q(\pi)$ with three more auxiliary variables. The new variational prior $q$ can be factorised as

\begin{equation}
    q({\bm\lambda}, \Omega, \Upsilon, F, \Pi, V, G, Z, T)=q({\bm\lambda})q(\Omega, \Upsilon)q(F)q(\Pi)q(V)q(G, Z)q(T),
\end{equation}
where the inference processes of $q(V), q(T)$ are identical to EBCC.

\subsection{Update rule for $q(\Omega,\Upsilon)$}
\label{apdx:omega and upsilon}
To get the optimal solution for $q(\Omega,\Upsilon)$, we need to find the exponential family form of its posterior. According to the PGM of \mname, we have
\begin{align*}
    \log q(\Omega, \Upsilon)
    &=\E_{\Pi, F, {\bm\lambda}}\left[\log\prod_{i=1}^N\prod_{k=1}^K\prod_{m=1}^M p(\pi_{ikm}\mid f_{ikm}, \upsilon_{ikm}, \omega_{ikm})p(\omega_{ikm}\mid\upsilon_{ikm})p(\upsilon_{ikm}\mid\lambda_{i})\right] \\
    &=\E_{\Pi,F,{\bm\lambda}}\Bigg[ \log \prod_{i=1}^N\prod_{k=1}^K\prod_{m=1}^M 2^{-(\pi_{ikm}+\upsilon_{ikm})}\exp\left\{ \frac{(\pi_{ikm}-\upsilon_{ikm})\sum_{i=1}^N f_{ikm}}{2} - \frac{(\sum_{i=1}^N f_{ikm})^2}{2}\omega_{ikm} \right\}\cdot \\
    &\ \ \ \ \PG(\omega_{ikm}\mid\upsilon_{ikm},0)\frac{\lambda_i^{\upsilon_{ikm}}\exp(-\lambda_i)}{\upsilon_{ikm}!} \Bigg] \\
    &=\E_{\Pi,F,{\bm\lambda}}\Bigg[ \sum_{i=1}^N\sum_{k=1}^K\sum_{m=1}^M\Bigg\{ -(\pi_{ikm}+\upsilon_{ikm})\log 2 + \frac{(\pi_{ikm}-\upsilon_{ikm})\sum_{i=1}^N f_{ikm}}{2} - \frac{(\sum_{i=1}^N f_{ikm})^2}{2}\omega_{ikm} \\
    &\ \ \ \ +\log\PG(\omega_{ikm}\mid\upsilon_{nk}, 0) + \upsilon_{ikm}[\psi(\alpha_i)-\psi(\beta_i)]-\log\upsilon_{ikm}!\Bigg\}\Bigg] \\
    &=\sum_{i=1}^N\sum_{k=1}^K\sum_{m=1}^M\Bigg\{ -(\E_{\Pi}[\pi_{ikm}] + \upsilon_{ikm})\log2 - \upsilon_{ikm}\frac{\E_F[\sum_{i=1}^N f_{ikm}]}{2} - \omega_{ikm}\frac{\E_F[ (\sum_{i=1}^N f_{ikm})^2]}{2} \\
    & \ \ \ \ +\log\PG(\omega_{ikm}\mid\upsilon_{ikm},0) + \upsilon_{ikm}[\psi(\alpha_i)-\log\beta_i]-\log\upsilon_{ikm}! \Bigg\}
\end{align*}
which implies that $q(\omega_{ikm}, \upsilon_{ikm})$ follows the distribution given below:
\begin{align*}
    q(\omega_{ikm}, \upsilon_{ikm})
    &\propto\left(\exp(-\frac{\widehat{m_{ikm}}}{2})\right)^{\upsilon_{ikm}}\exp\left(-\frac{(\mean f_{ikm})^2}{2}\omega_{ikm}\right)\PG(\omega_{ikm}\mid\upsilon_{ikm},0)\left(\frac{\exp(\psi(\alpha_i))}{\beta_i}\right)^{\upsilon_{ikm}}\frac{1}{\upsilon_{ikm}!} \\
    &\propto\left\{\exp\left(-\frac{(\mean f_{ikm})^2}{2}\omega_{ikm}\right)\PG(\omega_{ikm}\mid\upsilon_{ikm},0)\cosh^{\upsilon_{ikm}}\left(-\frac{(\mean f_{ikm})^2}{2}\right)\right\} \\
    &\propto\PG(\omega_{ikm}\mid\upsilon_{ikm}, c_{ikm})\Po(\upsilon_{ikm}\mid\gamma_{ikm}),
\end{align*}
where
\begin{align}
    c_{ikm} &= \mean f_{ikm}=\sqrt{\widehat{m_{ikm}}^2 + \widehat{\Sigma_k}(i,i)} \label{eq:c}\\
    \gamma_{ikm} &= \frac{\exp(\psi(a_i))\exp(-\frac{\widehat{m_{ikm}}}{2})}{\beta_i\cosh(\frac{\mean f_{ikm}}{2})} \label{eq:gamma}
\end{align}

\subsection{Update rule for $q(\lambda)$}
\begin{align*}
    \log q({\bm\lambda})
    &=\E_\Upsilon[\log p({\bm\upsilon\mid\lambda})p({\bm\lambda})] \\
    &=\sum_{i=1}^N\left[\sum_{k=1}^K\sum_{m=1}^M\gamma_{ikm}\log\lambda_i-K\lambda_i\right]
\end{align*}
where computation of $\gamma_{ikm}$ is given by Eq.~\ref{eq:gamma}. This implies that $q(\lambda_i)\sim\operatorname{Ga}(\lambda_i\mid a_i,b_i)$ where
\begin{align}
    a_i &= \sum_{k=1}^K\sum_{m=1}^M\gamma_{ikm} + 1 \\
    b_i  &= K
\end{align}

\subsection{New update rule for $q(\Pi)$}
\begin{align*}
    \log q(\Pi) 
    &= \E_{F,\Upsilon, \Omega, Z, G}\left[ \log\prod_{i=1}^N\prod_{k=1}^K\prod_{m=1}^M p(\pi_{ikm}\mid f_{ikm}, \upsilon_{ikm}, \omega_{ikm}) p(g_i=m\mid\pi_{ikm}, z_i) \right] \\
    &= \E_{F,\Upsilon, \Omega, Z, G}\Bigg[\sum_{i=1}^N\sum_{k=1}^K\sum_{m=1}^M - (\pi_{ikm}+\upsilon_{ikm})\log2 + \frac{(\pi_{ikm}-\upsilon_{ikm})f_{ikm}}{2}-\frac{(f_{ikm})^2}{2}\omega_{ikm} \\
    & \ \ \ \ + q(z_i=k, g_i=m)\log\pi_{ikm}  \Bigg] \\
    &= \sum_{i=1}^N\sum_{k=1}^K\sum_{m=1}^M \E_{Z, G}[q(z_i=k, g_i=m)]\log\pi_{ikm}-(\log2)\pi_{ikm} + \frac{\E_F[f_{ikm}]}{2}\pi_{ikm} + \co,
\end{align*}
which implies that $q(\pi_{ikm})$ follows the distribution given below:
\begin{align*}
    q(\pi_{ikm})
    &\propto \pi_{ikm}^{\rho_{ikm}}\cdot e^{ -(\log2 - \frac{\hat{m}_{km}}{2})\pi_{ikm}} \\
    &\propto \ga(\pi_{ikm}\mid \phi_{ikm}, \xi_{ikm}),
\end{align*}
where 
\begin{align*}
    \phi_{ikm} &= \rho_{ikm} + 1 \\
    \xi_{ikm} &= \log2-\frac{\hat{m}_{ikm}}{2}
\end{align*}

\subsection{Update rule for $q(F)$}
\begin{align*}
    \log q(F) 
    &= \E_{\Pi, \Upsilon, \Omega}\left[ \log\prod_{i=1}^N\prod_{k=1}^K p(\pi_{ikm}\mid f_{ikm}, \upsilon_{ikm}, \omega_{ikm})p(f_{ikm}) \right] \\
    &=\E_{\Pi, \Upsilon, \Omega}\left[\log\prod_{k=1}^K\N(f_{ikm}\mid\frac{{\pi}_{km}-{\bm\upsilon}_{km}}{2}, \diag({\bm\omega}_{km})^{-1})\N(f_{ikm}\mid{\bm 0}, \Sigma_{km})\right]
\end{align*}
which implies that $q({\bm f}^{(k, m)})\sim\N({\bm f}^{(k, m)}\mid\hat{\bm m}_k, \hat{\Sigma}_k)$ where
\begin{align}
    \hat{\bm m}_{km} &= \frac{1}{2}\hat{\bm\Sigma}_{km}(\bm{\phi}_{km}/\bm{\xi}_{km}-\E[{\bm \upsilon_{km}}]) \\
    \hat{\bm \Sigma}_{km} &= ({\bm\Sigma}_{km}^{-1}+\diag(\E[\hat{\bm \omega}_{km}]))^{-1}.
\end{align}
Note that $\E[{\bm \upsilon_{km}}]={\bm \gamma}_{km}, \E_{q(\omega_{km}, \upsilon_{km})}[\omega_{km}]=\frac{\E[{\bm \pi_{ikm}}]+{\bm \gamma_{km}}}{2{\bm c_{km}}}\tanh\frac{{\bm c_{km}}}{2}$.
\end{document}